# Overview of Class Activation Maps for Visualization Explainability

Pham Thi Minh Anh
Queen Mary University of London
t.pham@se22.qmul.ac.uk

***Abstract*— Recent research in deep learning methodology has led to a variety of complex modelling techniques in computer vision (CV) that reach or even outperform human performance. Although these black-box deep learning models have obtained astounding results, they are limited in their interpretability and transparency which are critical to take learning machines to the next step to include them in sensitive decision-support systems involving human supervision. Hence, the development of explainable techniques for computer vision (XCV) has recently attracted increasing attention. In the realm of XCV, Class Activation Maps (CAMs) have become widely recognized and utilized for enhancing interpretability and insights into the decision-making process of deep learning models. This work presents a comprehensive overview of the evolution of Class Activation Map methods over time. It also explores the metrics used for evaluating CAMs and introduces auxiliary techniques to improve the saliency of these methods. The overview concludes by proposing potential avenues for future research in this evolving field.**

## I. Introduction

In recent years, the research community has witnessed a growing interest in explaining neural network predictions, as it can enhance the transparency of learned complex models and enable the justification of incorrect outputs in a human-comprehensible way. While various attempts have been made to offer explanations in different forms [30, 31, 32, 35, 36], including textual or symbolic representations, the graphical visualization of key aspects, such as regions of input data, has emerged as one of the most straightforward and effective approaches to providing insights into neural network inferences [33, 34].

Originally introduced by Zhou et al. [1], the concept behind Class Activation Maps (CAMs) involves mapping out the regions within an image that the CNN deems most pertinent for recognizing a particular category. The motivation of this approach arises from the distinct spatial information within each activation map, with closer convolutional layers to the network's classification stage offering more effective high-level activations for visual localization, thereby explaining the final prediction. Zhou et al devised a method to project the weights of the CNN's output layer back onto the feature maps generated by the convolutional layers. This projection effectively creates a heatmap, illuminating the areas within the image that significantly influence the network's decision-making process. As CAMs gained recognition and popularity in the computer vision community, researchers and practitioners explored various avenues to further improve and diversify CAM methods [2]. In this review, we will summarize the recent developments, present different methods of CAM that have been proposed in the context of deep neural networks, provide evaluation metrics, and highlight the current best practices when applying these methods. We also propose potential directions of study.

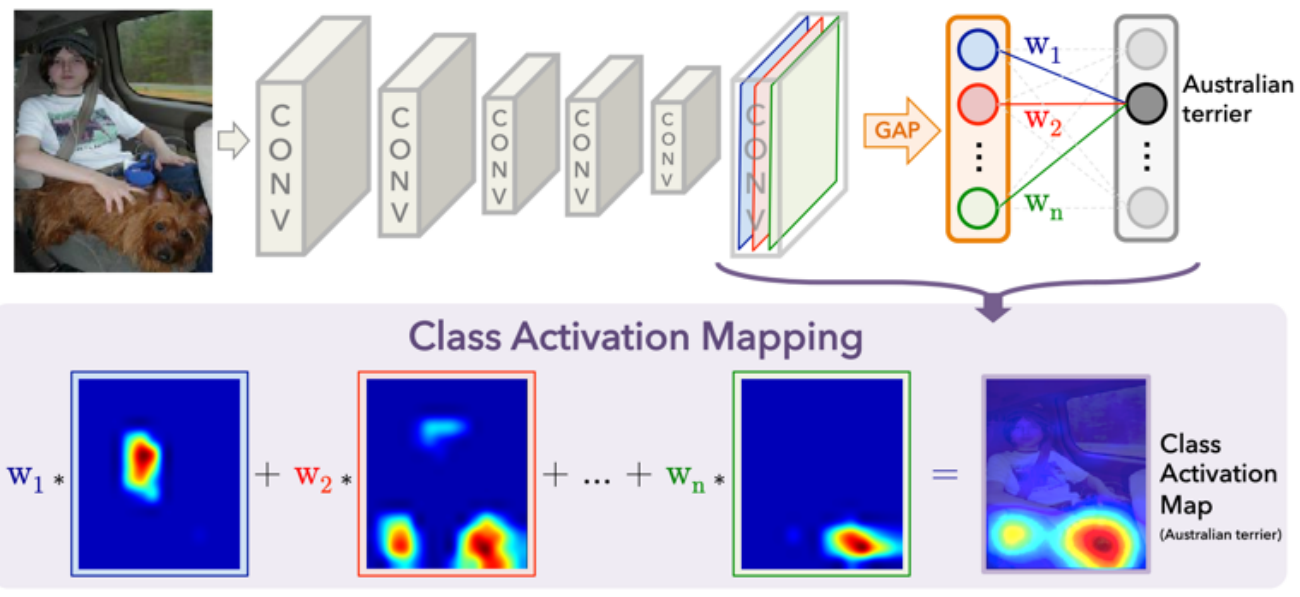

Fig. 1. Overview of CAM approaches for explaining predictions: explanation maps are produced via a linear combination of the activations of a convolutional layer. Taken from [1].

## II. Practical CAM Methods

In this section, we focus on two main categories of CAM techniques: Gradient-Based CAM Methods [3, 4, 18, 19, 20] and Gradient-Free CAM Methods [33, 22,19, 18]. In our view, these techniques exemplify the current diversity of possible CAM-based visualization approaches to explaining predictions in terms of input features, and, taken together, provide broad coverage of the types of CAM models to explain the practical computer vision tasks.

### *A. Class Activation Map*

Zhou et al. [1] introduced the original CAM method, which generates class-discriminative visualization maps. This is achieved by linearly combining activation maps from the final convolutional layer with importance coefficients, represented by the Fully Connected (FC) weights associated with the target class. In the context of the final convolutional layer's output feature map, denoted as $A$, comprising a set of activation maps from $A_1$ to $A_N$, the Class Activation Map is defined as follows:

$$L^c_{CAM} = \sum_{n=1}^{N} w^c_n A^n \tag{1}$$

where $w^c_n$ represents the weight associated with class c from the Fully Connected layer functioning as a classifier. So CAM is constrained by the model architecture, necessitating the presence of two specific components: a Global Average Pooling (GAP) layer and one Fully Connected layer that serves as its classifier.

### *B. Gradient-Based Methods*

Gradient-based methods leverage gradients computed during backpropagation to generate heatmaps that highlight the discriminative regions within an image for specific categories as identified by Convolutional Neural Networks (CNNs). As GAP layers were only present in select neural networks, Grad-CAM [3] extends the original CAM method by allowing for the use of gradients from any target category flowing into the final convolutional layer, making it applicable to a broader range of CNN architectures. Grad-CAM is formulated by the following,

$$L^c_{G-CAM} = \text{ReLU}\left( \sum_{n=1}^{N} a_n A^n \right) \tag{2}$$

where the importance coefficients are computed as follows:

$$a_n = \frac{1}{Z}\sum_i \sum_j \frac{\partial y^c}{\partial A_{i,j}^n} \quad (3)$$

Here, Z is the number of pixels in the feature map $A^n$, $y^c$ is the classification score for class $c$, and $A_{i,j}^n$ refers to the activation value at location $(i,j)$ on $A^n$. Grad-CAM++ [4] further refines this approach by calculating the true weighted average of gradients, enhancing the precision and reliability of heatmap generation. Following the same principle of using the partial derivative $\frac{\partial y^c}{\partial A_{i,j}^n}$ as Grad-CAM, later methods such as Layer-CAM [18] and XGrad-CAM [19] introduced new variations in gradient computations and improvements on heatmap visualization process. Each method diverges in its approach to calculating gradients, all with the shared goal of refining the precision and stability of the gradient-based visualization method, thereby offering researchers and practitioners a spectrum of options to choose from based on their specific needs and objectives in interpreting deep neural network decisions.

However, these methods rely on trainable models to leverage gradient information, which can limit their applicability in post-deployment frameworks such as ONNX [23] or OpenVINO [24]. Moreover, the gradients of deep neural networks tend to diminish due to the gradient saturation problem. There is also the case of gradients generating false confidence [5], where the highest weight of the activation map does not lead to the largest increase in confidence.

### C. Gradient-Free Methods

While Gradient-Based CAM Methods have been prominent, there has been a growing interest in developing alternative approaches that do not rely on gradients. Wang et al., [5] proposed Score-CAM using the weight of each activation map through its forward passing score on the target class, to eliminate the dependence on gradients. They also propose a "Channel-wise Increase in Confidence", a normalized mask on the top of the input to obtain the feature of interest in the input:

$$L_{Score-CAM}^c = \text{ReLU}\left(\sum_{n=1}^N softmax(F^c(X') - F^c(X))A^n\right) \quad (4)$$

Where $F^c$ and $X$ denote a CNN model and an input image respectively, and $X'$is Hadamard Product of $X$ and $norm(up(A^n))$. The function up $(\cdot)$ represents the up-sampling operation, which scales the activation map An to match the dimensions of the original input image $X$.

Recipro-CAM [6] takes the approach of occluding all activations except for the target pixel that is mapped to the region of the original input in terms of the receptive field. The target layer's feature map is modified by multiplying it with a set of HW spatial masks, each containing a single value corresponding to a specific position in the feature map. This multiplication generates a fresh set of HW feature maps, which are subsequently inputted into the next stage of the neural network. The predicted scores corresponding to the target class are gathered and utilized to fill each position within the resulting saliency map. Here, K represents the number of channels, while H and W signify the feature height and width. This process allows them to highlight the importance of each activation on the output feature map. It is currently state-of-the-art in terms of the ADCC metric [7] and in terms of computational efficiency. Fig 2. portrays a visual comparison of the methods discussed above.

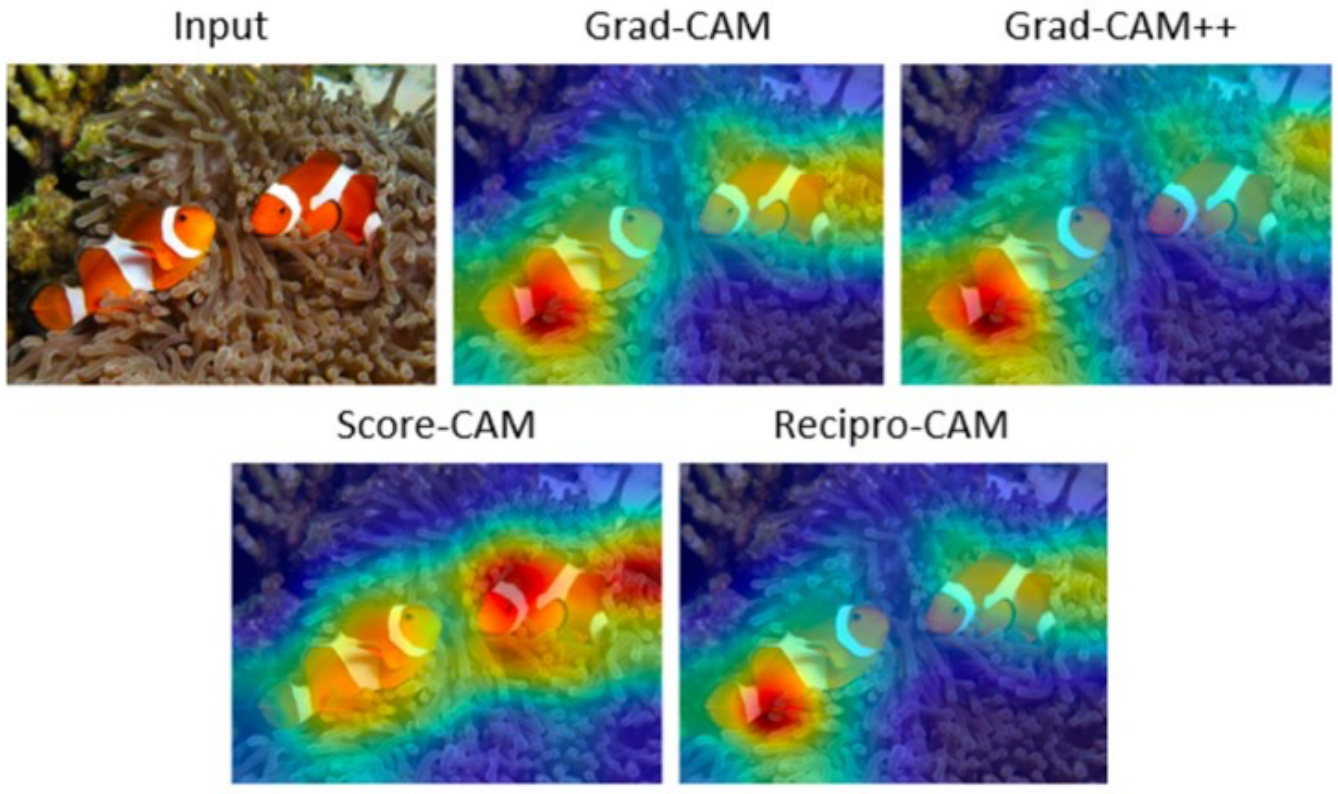

Fig. 2. Visual depiction of Heatmaps. Taken from [6].

## III. Combining Smoothing and Integration Functions

Smilkov et al., [8] identified noise in sensitivity maps as meaningless local variations in partial derivatives. They introduced the smoothing method to control noise during the generation of gradients by adding sufficient noise to the input data and generating similar images. Then the average of the resulting sensitivity maps of each of these sampled, similar images are taken. By averaging these noisy gradients over multiple iterations, they generate a more stable gradient map that effectively reduces the noise present in the original gradient. The visually-sharp gradient-based sensitivity map is indicated by:

$$M'_c(x) = \frac{1}{N}\sum_{n=1}^N M_c(x + \mu(0, \sigma^2))). \quad (5)$$

where $N$ is the number of similar samples from input $x$ which are created by adding the Gaussian noise with a standard deviation σ. This technique can be applied to various deep-learning interpretability methods to enhance their quality and mitigate artifacts caused by noise in the gradients. Since then, Grad-CAM and Grad-CAM++ have been improved through SmoothGrad [8] and Smoothed-Grad-CAM++ [10]. In the line of gradient-free CAM methods, Wang et al. [12] also integrated smoothing into the pipeline of ScoreCAM to generate smoother feature localization method SS-CAM by using a smoothness regularization on the activation map.

Besides the noise issue on visualization maps, Sundarajan et al., [9] identified that previous methods lacked adaption to sensitivity and implementation invariance. A significant limitation when implementing backpropagation in attribution methods is their violation of the sensitivity axiom. This axiom implies that for any given pair of input and baseline images that differ in just one feature, an attribution method should distinctly emphasize this difference by assigning varying values to that particular feature. The second axiom, implementation invariance, means that two networks are functionally equivalent if their outputs are equal for all inputs, despite having very different implementations. To solve these issues, [9] introduced integrating gradients over the CAM pipelines. Integrated Gradients is a technique that enhances the justification of the sensitivity of output confidence scores to input features. It achieves this by calculating the integral of gradient values along any continuous path connecting a predefined baseline and the input image. This approach provides a more thorough understanding of how the model's predictions are influenced by various input features along the entire path, thereby improving the interpretability of the model's behaviour. Since then, Grad-CAM has been improved through Integrated Grad-CAM [11] by replacing the gradient terms in Grad-CAM with similar terms based on Integrated Gradient. Score-CAM has also been further improved via IS-CAM [13].

## IV. Helper Methods to Improve Saliency

In recent developments, efforts have been made to enhance the generation of saliency maps within class activation maps (CAMs). Oh et al. [14] have introduced a technique that leverages image transformations, based on the idea that critical regions containing more informative features result in larger transformations. This approach not only improves the quality of saliency maps but also reduces the associated computational costs, making it more efficient.

Another approach, proposed by Tursun et al. [15], is a model-agnostic extension. It involves the fusion of saliency maps extracted from multiple patches at various scales and different regions of an image. This fusion is accomplished through the use of channel-wise weights and spatially weighted averages, resulting in more comprehensive and informative saliency maps that capture the most relevant features across different image areas and scales.

## V. Metric Evaluation

In an ideal scenario, a CAM-based explanation map should encompass the smallest possible set of pixels that are essential for explaining the network's output. While this aspect has predominantly been assessed qualitatively, the quantitative evaluation of explanation capabilities is still in its earliest stages, marked by the emergence of various evaluation metrics [4, 16, 19].

### *A. Qualitative Analysis*

Qualitative evaluation for CAM-based methods typically involves visually assessing the generated saliency maps or heatmaps. This evaluation primarily revolves around visual inspection and employs several key criteria. Firstly, localization accuracy is paramount, aiming to gauge the method's ability to precisely identify and highlight relevant objects or regions within an input image. Clarity and consistency are equally vital, ensuring that the explanation map is lucid, devoid of extraneous noise or irregularities, and consistently provides coherent results across similar inputs. Robustness evaluation examines the method's performance across diverse input images and object variations, seeking to confirm its reliability in consistently producing meaningful and accurate heatmaps. While qualitative evaluation is essential for assessing the visual quality and interpretability of CAM-based explanation maps, it is often complemented by quantitative evaluation metrics to provide a more comprehensive assessment of model performance.

### *B. Quantitative Analysis*

Quantitative analysis of CAM-based methods encompasses two essential aspects: localization capacity evaluation and faithfulness evaluation. Localization capacity assesses the method's ability to accurately pinpoint discriminative regions within input images. Metrics used to measure localization capacity include the energy-based point game proportion [5, 25] and intersection over union (IoU) [1, 3, 18]. These metrics provide insights into how precisely the explanation map can identify and highlight the regions that significantly contribute to the model's decision. Faithfulness evaluation, on the other hand, focuses on how well the explanation map faithfully represents the model's decision process, ensuring that it accurately reflects the model's inner workings. The energy-based point game proportion is a metric used to quantify the extent to which the energy (or saliency) within the explanation map aligns with the bounding box of the target object in the image. It is formulated as follows:

$$Proportion = \frac{\sum_{i,j \in bbox}(norm(up(L)))_{(i,j)}}{\sum_{i,j \in \Lambda}(norm(up(L)))_{(i,j)}} \quad (6)$$

where Λ represents the size of the original image, bbox represents the ground truth bounding box of the target object, and (i,j) denotes the location of a pixel within the image. Intersection over Union (IoU) is a metric commonly used in object localization tasks to evaluate the overlap between an estimated bounding box and a ground truth bounding box. To generate an estimated bounding box from the CAM-based methods, we first binarize the saliency map with the threshold of the max value of the saliency map. Once a bounding box is created based on the most significant connected segments of pixels in the binarized map, IoU is calculated by the following:

$$IoU = \frac{\sum_{i,j \in (bbox_e \cap bbox_g)} 1}{\sum_{i,j \in (bbox_e \cup bbox_g)} 1} \quad (7)$$

where $bbox_e$ represents the estimated bounding box while $bbox_g$ represents the ground truth bounding box.

The second aspect, faithfulness measures, focuses on determining the significance and accuracy of the regions highlighted by the explanation map in relation to the model's decision. The goal is to ensure that the explanation map faithfully represents the model's internal decision-making process. Therefore, the faithfulness evaluation has been widely performed for interpretable CAM-based visualization methods [3, 4, 5, 25, 26, 27]. To evaluate the faithfulness of the interpretable visualization methods, Chattopadhay et al., [4] proposed averaged drop (AD) and increase in confidence (IC) metrics and Petsuik et al., [16] proposed causal metrics: deletion and insertion curves.

**Average Drop.** This metric focuses on measuring the impact of showing only the explanation map, rather than the full input image, on the model's confidence in predicting the target class. The Average Drop is calculated as follows:

$$AD = \frac{1}{N}\sum_{n=1}^{N}\frac{\max(0, s_i - s_i')}{s_i} \times 100 \quad (8)$$

Here, $s_i$ is the model's confidence score for the target class c when using the full image and $s_i'$ is the model's confidence score for the target class c when it sees only the explanation map for the i-th instance.

**Increase in Confidence.** It assesses how often the explanation map leads to increased confidence in the model's predictions by counting the number of instances in which the model's confidence for a particular class is greater when using the explanation map than when using the full input image. The value is then averaged over different images:

$$IC = \frac{1}{N}\sum_{n=1}^{N} 1_{s_i < s_i'} \times 100 \quad (9)$$

TABLE I.
Comparison of Class Activation Map approaches considered for the study.

| CAM Models | Model Details | | | Evaluation Metric |
|---|---|---|---|---|
| | *Methodology* | *Merits* | *Limitations* | *ADCC-Score** |
| **CAM [1]** | Linear combination of GAP on last conv layer feature map | - Pilot study in CAM.<br>- Took advantage of GAP Layer merits. | GAP Layers are not present in all networks. | - |
| **Grad-CAM [3]** | Gradient of target class score with respect to activation map. | Generalised CAM. | Gradient based methods succumb to gradient explosion and false confidence. | 76.4 |
| **Grad-CAM ++ [4]** | True weighted average of the gradients. | - Generalised Grad-CAM.<br>- Proposed new evaluation Metrics. | Gradient based methods succumb to gradient explosion and false confidence. | 76.34 |
| **Score-CAM [5]** | Weight of each activation map through its forward passing score on target class. | - Overcame gradient issues.<br>- Paved way for score based approaches. | Computationally too expensive. | 78.55 |
| **Recipro-CAM [6]** | Prediction scores of each class for feature location imposed by each of the new input feature map's spatial masks. | Current State of the art. | Not reproducible. | 81.38 |

*Values based on Recipro-CAM Reports

**Insertion and Deletion.** It is based on the change in decision forced on the model when pixels are either removed or added to the original decision [16]. These two metrics calculate the decrease and increase in the ability of the CAM to function when occlusions are introduced to it. Both Deletion and Insertion metrics are typically expressed in terms of the total Area Under the Curve (AUC), which provides a comprehensive summary of how the model's prediction confidence changes with the gradual modification of the image content.

However, Poppi et al., [7] clearly demonstrated that AD, IC, Insertion and Deletion metrics can be easily tampered with to produce artificially high results. To counter this, the authors proposed the Average DCC metric (ADCC). It takes into consideration multiple aspects, including the average drop in model confidence (Average Drop), the coherence of the explanation (Maximum Coherence), and the complexity of the explanation (Complexity). Coherency focuses on containing all relevant features and complexity focuses on the minimum set to do so while Complexity evaluates the simplicity and interpretability of the explanation. By combining these aspects into a single score, ADCC provides a holistic assessment of the quality and effectiveness of an explanation method. The calculation of ADCC involves computing the harmonic mean of these components:

$$ADCC = 3\left(\frac{1}{Coherency(x)} + \frac{1}{1 - Complexity(x)} + \frac{1}{1 - AD(x)}\right)^{-1} \quad (10)$$

Table 1. Depicts a comparison of some mentioned CAM-based methods in the overview with respect to ADCC.

### C. Performance Analysis

Previous CAM-based XAI research did not consider execution performance. This was primarily due to the fact that CAM and gradient-based CAM methods were typically fast and efficient enough, rendering performance issues as a non-significant factor. However, in more recent developments, emerging gradient-free methods like Score-CAM [5] and Ablation-CAM [22] have brought performance concerns to the forefront. The performance of these newer methods is intricately tied to factors such as the number of feature channels, input resolution, and the capacity of the neural network being used. These methods may exhibit performance bottlenecks or inefficiencies under certain conditions, necessitating a reevaluation of execution speed and resource requirements in the context of gradient-free CAM-based XAI research.

## VI. Discussion and Future Work

### A. Diagnosing image classification CNNs

Grad-CAM introduced descriptive methods to:

- Understanding Classification Errors: Grad-CAM provides insights into the specific regions of an image that influence the network's incorrect classification.
- Bias Identification and Mitigation: It helps in uncovering whether the model is overly focusing on certain image features, which might result in biased predictions.

Quantifying these steps and establishing an empirical process to quantify these descriptions can lead to interesting strides in image classification pipelines.

### B. Exploring intensity transformations in EVET [14]

EVET discussed significant improvements through introducing 4 geometrical transformations. Integrating intensity transformations into the EVET pipeline can potentially enhance the capabilities of this system for providing visual explanations of deep neural networks. Intensity transformations can encompass various techniques, including contrast adjustments, histogram equalization, and adaptive filtering, among others. These transformations may highlight important features and patterns in the data that contribute to the model's prediction, thereby leading to more informative and interpretable visual explanations of deep neural network decisions. It will be interesting to see how the pipeline fares with intensity transformations.

### C. Experiment with different image settings.

A major advancement in Grad-CAM ++ over Grad-CAM was the ability to improve performance in the cases of multiple occurrences of the same image. This research direction involves conducting experiments with diverse image conditions and scenarios to gain deeper insights into the strengths and limitations of various Class Activation Map (CAM) techniques. A specialized dataset should be created containing various scenarios such as large objects, small objects, multiple "same" objects, multiple "different" objects and more, to develop intricate understandings of where different CAMs falter. It will also prove to be helpful in terms of how different networks respond to these different data settings.

## D. Further research directions

Exploring multi-label scenarios, where images may contain multiple objects or concepts, is an important extension of this research direction. Understanding how CAMs can adapt to scenarios with multiple labels and potentially overlapping regions of interest is a valuable pursuit. Furthermore, the field could explore how neural networks function and interact with CAMs. This involves delving into the inner workings of these models to uncover the nuances of their decision-making processes when guided by CAM-based explanations. By aligning research efforts with practical applications and enhancing our comprehension of network behaviour, the CAM community can contribute to the advancement of Explainable Artificial Intelligence and facilitate the integration of CAMs into a wide range of domains where interpretability and transparency are crucial.

# VI. Conclusion

This overview provides important information about how CAMs have developed over the years to solve different drawbacks associated with them. With the establishment of robust evaluation metrics and the development of relatively high-performing models, the future trajectory of CAM-based methods should shift towards more application-oriented research [17]. Rather than solely focusing on general improvements over existing approaches, researchers could aim to explore how CAMs can be effectively applied in real-world scenarios. This includes investigating specific use cases and domains where CAMs can provide meaningful interpretability and insights.

# References

[1] Zhou, Bolei, et al. "Learning deep features for discriminative localization." *Proceedings of the IEEE conference on computer vision and pattern recognition*. 2016.

[2] Lin, Min, Qiang Chen, and Shuicheng Yan. "Network in network." *arXiv preprint arXiv:1312.4400* (2013).

[3] Selvaraju, Ramprasaath R., et al. "Grad-cam: Visual explanations from deep networks via gradient-based localization." *Proceedings of the IEEE international conference on computer vision*. 2017.

[4] Chattopadhay, Aditya, et al. "Grad-cam++: Generalized gradient-based visual explanations for deep convolutional networks." *2018 IEEE winter conference on applications of computer vision (WACV)*. IEEE, 2018.

[5] Wang, Haofan, et al. "Score-CAM: Score-weighted visual explanations for convolutional neural networks." *Proceedings of the IEEE/CVF conference on computer vision and pattern recognition workshops*. 2020.

[6] Byun, Seok-Yong, and Wonju Lee. "Recipro-CAM: Gradient-free reciprocal class activation map." *arXiv preprint arXiv:2209.14074* (2022).

[7] Poppi, Samuele, et al. "Revisiting the evaluation of class activation mapping for explainability: A novel metric and experimental analysis." *Proceedings of the IEEE/CVF Conference on Computer Vision and Pattern Recognition*. 2021.

[8] Smilkov, Daniel, et al. "Smoothgrad: removing noise by adding noise." *arXiv preprint arXiv:1706.03825* (2017).

[9] Sundararajan, Mukund, Ankur Taly, and Qiqi Yan. "Axiomatic attribution for deep networks." *International conference on machine learning*. PMLR, 2017.

[10] Omeiza, Daniel, et al. "Smooth grad-cam++: An enhanced inference level visualization technique for deep convolutional neural network models." *arXiv preprint arXiv:1908.01224*(2019).

[11] Sattarzadeh, Sam, et al. "Integrated grad-CAM: Sensitivity-aware visual explanation of deep convolutional networks via integrated gradient-based scoring." *ICASSP 2021-2021 IEEE International Conference on Acoustics, Speech and Signal Processing (ICASSP)*. IEEE, 2021.

[12] Wang, Haofan, et al. "SS-CAM: Smoothed Score-CAM for sharper visual feature localization." *arXiv preprint arXiv:2006.14255*(2020).

[13] Naidu, Rakshit, et al. "IS-CAM: Integrated Score-CAM for axiomatic- based explanations." *arXiv preprint arXiv:2010.03023* (2020).

[14] Oh, Youngrock, et al. "EVET: enhancing visual explanations of deep neural networks using image transformations." *Proceedings of the IEEE/CVF Winter Conference on Applications of Computer Vision*. 2021.

[15] Tursun, Osman, et al. "SESS: Saliency Enhancing with Scaling and Sliding." *European Conference on Computer Vision*. Springer, Cham, 2022.

[16] Petsiuk, Vitali, Abir Das, and Kate Saenko. "Rise: Randomized input sampling for explanation of black-box models." *arXiv preprint arXiv:1806.07421* (2018).

[17] Saporta, A., Gui, X., Agrawal, A. *et al.* Benchmarking saliency methods for chest X-ray interpretation. *Nat Mach Intell* **4**, 867–878 (2022).

[18] P. Jiang, C. Zhang, Q. Hou, M. Cheng, and Y. Wei, ”LayerCAM: Exploring hierarchical class activation maps for localization,” IEEE Trans Image Process., vol. 30, pp. 5875–5888, 2021.

[19] R. Fu, Q. Hu, X. Dong, Y. Guo, Y. Gao, and B. Li, ”Axiom-based gradCAM: Towards accurate visualization and explanation of CNNs,” 2020, arXiv:2008.02312.

[20] Daniel Omeiza, Skyler Speakman, Celia Cintas, and Komminist Weldermariam. Smooth grad-cam++: An enhanced inference level visualization technique for deep convolutional neural network models. arXiv preprint arXiv:1908.01224, 2019.

[21] Daniel Smilkov, Nikhil Thorat, Been Kim, Fernanda Viégas, and Martin Wattenberg. Smoothgrad: removing noise by adding noise. arXiv preprint arXiv:1706.03825, 2017.

[22] Saurabh Desai and Harish G Ramaswamy. Ablation-cam: Visual explanations for deep convolutional network via gradient-free localization. In WACV, pages 972–980, 2020.

[23] J. Bai, F. Lu, K. Zhang, et al. ONNX: Open neural network exchange, 2019. URL https://github.com/onnx/onnx.

[24] Intel. OpenVINO toolkit, 2019. URL https://software.intel.com/enus/openvino-toolkit.

[25] H. Jung and Y. Oh, ”Towards better explanations of class activation mapping,” in Proc. IEEE/CVF Int. Conf. Comput. Vis. (ICCV), Oct. 2021, pp. 1366–1344.

[26] K. H. Lee et al. ”LFI-CAM: Learning feature importance for better visual explanation” in Proc. IEEE/CVF Conf. Comput. Vis. Pattern Recognit. (CVPR), Jun. 2021, pp. 1355—1363.

[27] J. R. Lee, S. Kim, I. Park, T. Eo, and D. Hwang, ”Relevance-CAM: Your model already knows where to look,” in Proc. IEEE/CVF Conf. Comput. Vis. Pattern Recognit. (CVPR), Jun. 2021, pp. 14944—14953

[28] M. B. Muhammad and M. Yeasin, ”Eigen-CAM: Class activation map using principal components,” in Proc. Int. Joint Conf. Neural Networks, 2020, pp. 1–7.

[29] R. Girshick, J. Donahue, T. Darrel, and J. Malik, ”Rich feature hierarchies for accurate object detection and semantic segmentation,” in Proc. IEEE/CVF Conf. Comput. Vis. Pattern Recognit. (CVPR), Jun. 2014, pp. 580–587.

[30] Yash Goyal, Ziyan Wu, Jan Ernst, Dhruv Batra, Devi Parikh, and Stefan Lee. Counterfactual visual explanations. In Proceedings of the International Conference on Machine Learning, 2019.

[31] Lisa Anne Hendricks, Zeynep Akata, Marcus Rohrbach, Jeff Donahue, Bernt Schiele, and Trevor Darrell. Generating visual explanations. In Proceedings of the European Conference on Computer Vision, 2016.

[32] Lisa Anne Hendricks, Ronghang Hu, Trevor Darrell, and Zeynep Akata. Grounding visual explanations. In Proceedings of the European Conference on Computer Vision, 2018.

[33] Karen Simonyan, Andrea Vedaldi, and Andrew Zisserman. Deep inside convolutional networks: Visualising image classification models and saliency maps. arXiv preprint arXiv:1312.6034, 2013.

[34] Matthew D Zeiler and Rob Fergus. Visualizing and understanding convolutional networks. In Proceedings of the European Conference on Computer Vision, 2014.

[35] Avanti Shrikumar, Peyton Greenside, and Anshul Kundaje. Learning important features through propagating activation differences. In Proceedings of the International Conference on Machine Learning, 2017.

[36] Jorg Wagner, Jan Mathias Kohler, Tobias Gindele, Leon Hetzel, Jakob Thaddaus Wiedemer, and Sven Behnke. Interpretable and fine-grained visual explanations for convolutional neural networks. In Proceedings of the IEEE/CVF Conference on Computer Vision and Pattern Recognition, 2019.